\documentclass{article}

\usepackage[main, preprint]{neurips_2026}

\usepackage[utf8]{inputenc}
\usepackage[T1]{fontenc}
\usepackage{hyperref}
\usepackage{url}
\usepackage{booktabs}
\usepackage{amsfonts}
\usepackage{nicefrac}
\usepackage{microtype}
\usepackage{xcolor}
\usepackage{graphicx}
\usepackage{subcaption}
\usepackage{amsmath}
\usepackage{amssymb}
\usepackage{mathtools}
\usepackage{amsthm}
\usepackage{bm}
\usepackage{algorithm}
\usepackage{algorithmic}

\theoremstyle{plain}
\newtheorem{theorem}{Theorem}[section]
\newtheorem{proposition}[theorem]{Proposition}

\newcommand{\Real}{\mathbb{R}}
\newcommand{\N}{\mathbb{N}}

\newcommand{\PS}{\mathrm{PS}}
\newcommand{\Matrix}[1]{\bm{\mathrm{#1}}}
\newcommand{\trans}{\intercal}
\newcommand{\diag}[2][]{\mathrm{diag}#1( #2 #1)}
\newcommand{\Norm}[3][]{#1\| #2 #1\|_{\mathrm{#3}}}
\newcommand{\One}{\Matrix{1}}
\newcommand{\Oh}[2][]{\mathcal{O}#1( #2 #1)}
\newcommand{\uh}{\mathrm{u}}
\newcommand{\set}[3][]{#1\{ #2 : #3 #1\}}

\title{LAMP: Look-Ahead Mixed-Precision Inference of Large Language Models}

\author{%
  Stanislav Budzinskiy \thanks{Corresponding author: \texttt{stanislav.budzinskiy@univie.ac.at}}\\
  Faculty of Mathematics\\
  University of Vienna, Austria\\
  \And
  Marian Gloser\\
  Faculty of Mathematics\\
  University of Vienna, Austria\\
  \And
  Tolunay Yilmaz\\
  Faculty of Mathematics\\
  University of Vienna, Austria\\
  \And
  Ying Hong Tham\\
  Huawei Technologies\\
  \And
  Yuanyi Lin\\
  Huawei Technologies\\
  \And
  Wenyi Fang\\
  Huawei Technologies\\
  \And
  Fan Wu\\
  Huawei Technologies\\
  \And
  Philipp Petersen\\
  Faculty of Mathematics\\
  University of Vienna, Austria\\
}

\begin{document}

\maketitle

\begin{abstract}
Mixed-precision computations are a hallmark of the current stage of AI, driving the progress in large language models towards efficient, locally deployable solutions.
This article addresses the floating-point computation of compositionally-rich functions, concentrating on transformer inference.
Based on the rounding error analysis of a composition $f(g(\Matrix{x}))$, we provide an adaptive strategy that selects a small subset of components of $g(\Matrix{x})$ to be computed more accurately while all other computations can be carried out with lower accuracy.
We then explain how this strategy can be applied to different compositions within a transformer and illustrate its overall effect on transformer inference.
We study the effectiveness of this algorithm numerically on GPT-2 models and demonstrate that already very low recomputation rates allow for improvements of up to two orders of magnitude in accuracy.
\end{abstract}


\section{Introduction}
Transformer deep neural networks (DNNs), originally introduced for sequence modeling in natural language processing \cite{vaswani2017attention}, have become a standard computational paradigm across a wide range of domains, including language understanding \cite{devlin2019bert} as well as vision and multimodal learning \cite{dosovitskiy2021image}.
From a computational perspective, a transformer is a deep composition of simple operators, obtained by iterating attention-based mappings and pointwise nonlinearities across layers. 

In practice, the evaluation of such deep compositions (\emph{inference}) is performed in floating-point (FP) arithmetic, low-precision formats being routinely employed to improve performance and energy efficiency \cite{gupta2015deep, micikevicius2018mixed, kalamkar2019study}.
From a numerical analysis standpoint, FP evaluation introduces rounding errors at every stage of the computation; the cumulative effect of these local errors hinges on how they propagate through successive compositions of operators \cite{higham2002accuracy}.

The bulk of operators in transformers are matrix products, and the existing approaches to mixed-precision inference largely address them based on two key principles: the input is quantized to low precision, and the output is accumulated in high precision \cite{xiao2023smoothquant}.
Typically, the quantization precision is uniform across the whole input, and the accumulation precision is uniform across the whole output.
While there are also mixed-precision quantization strategies \cite{dettmers2022gpt3}, we are not aware of developments in \emph{mixed-precision accumulation} for transformer inference.

In this work, we address this gap by proposing a mathematically principled mixed-precision inference strategy that is explicitly aware of compositional effects.
Instead of treating the output of an intermediate computation uniformly, we follow a \emph{look-ahead strategy} by flagging and recomputing with higher accuracy those computations whose round-off errors will be most strongly amplified by the ensuing operator. 
Our method has a rigorous theoretical foundation and delivers strong empirical results in numerical experiments, allowing for improvements of up to two orders of magnitude in accuracy with only 1\% of recomputations.
We describe our contribution in detail in Subsection~\ref{sec:contributions}.

\subsection{Rounding error analysis}
To place our contribution in context, we briefly review how rounding error analysis has been applied to function evaluation, and why DNNs fall outside this classical setting.
The primary applications are to matrix computations \cite{higham2002accuracy, connolly2021stochastic} and ``basic'' nonlinear functions such as the \emph{elementary functions}: powers and roots, exponentials and logarithms, trigonometric and hyperbolic functions \cite{muller2016elementary}. 

\emph{Special functions} \cite{gil2007numerical} are ``basic'' in mathematical physics, defined as solutions to specific differential equations or integrals.
The complexity of their evaluation stems from the need to discretize the differential equation or integral to sufficient (typically very high) precision, after which rounding effects of FP arithmetic become noticeable \cite{lauter2015semi}.

DNNs, including transformers, can be considered ``basic'' in the field of AI, and they differ from the aforementioned ``basic'' functions in two aspects: the target accuracy\footnote{How many unit round-offs of error are tolerable, rather than the value of the unit round-off.} and the source of evaluation complexity.
First, the applications of DNNs do not seem to require the accuracy to be very high.
Second, the evaluation complexity of DNNs is due to their extremely rich \emph{compositional structure}, where the ``building blocks'' are either elementary functions or their compositions \cite{blanchard2021accurately, el2024bounds}.

Therefore, \emph{the algorithms of DNN inference need to pay specific attention to compositions}.
Rounding error analysis applied to DNNs leads to worst-case bounds that grow exponentially with depth \cite{el2026mixed, budzinskiy2025numerical}.
While the \emph{global} rounding error appears to be difficult to tame, the \emph{local} rounding error at each composition is easier to control---the core idea of our work.

\subsection{Contributions, limitations, and outlook}\label{sec:contributions}
Our key contribution is methodological: we identify \emph{a new paradigm for mixed-precision evaluation in transformer inference}, placing it in the accumulation of matrix products and thereby complementing the standard approach to mixed-precision matrix multiplication rooted in the quantization of operands.
The attainable gains within this novel framework are validated in numerical experiments, the results of which simultaneously serve as a proof of concept for our idea and a benchmark for future practical inference kernels designed in accordance with it.
In addition to empirical accuracy improvements, our results highlight the importance of compositional numerical effects in deep models.

\textbf{Novel computational framework:}
We propose an adaptive approach to FP evaluation of compositions called \emph{look-ahead mixed precision} (LAMP).
This method stems from a rigorous theoretical derivation and is formulated explicitly as an optimization problem \eqref{eq:lamp_problem}, which aims to determine a sparse subset of the inner function's components that need to be recomputed more accurately to ensure the numerical stability of the composition.
Find the details in~Section~\ref{sec:theory}.

\textbf{Particularly well-suited to transformers:}
A theoretical investigation of the nonlinearities inherent to the transformer architecture allows us to prove that the aforementioned \emph{selection procedure admits closed-form solutions}, overcoming the otherwise combinatorial nature of the problem and paving the way for practical implementations.
See Section~\ref{sec:theory_llm}.

\begin{figure}[h!]
    \centering
    \includegraphics[width=0.65\linewidth]{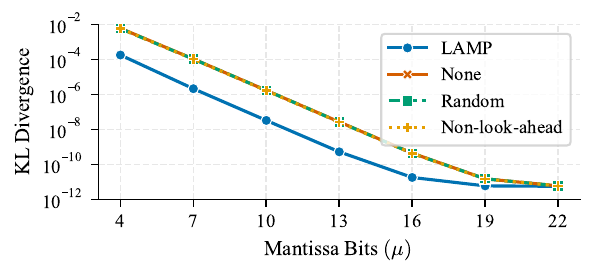}
    
    \caption{Performance of mixed-precision GPT-2 XL inference on the OpenWebText dataset with the proposed LAMP evaluation of KQ inner products.
    Only 0.9\% of them are recomputed in FP32.
    The Kullback--Leibler divergence is measured against a reference model with uniform FP32 accumulation.}
    \label{fig:bits_lamp_vs_fake}
\end{figure}

\textbf{Proof of concept:}
For validation, we apply LAMP to evaluate softmax in attention.
Our method demonstrates convincing performance in numerical experiments with the GPT-2 XL model.
In Figure~\ref{fig:bits_lamp_vs_fake}, the key-query (KQ) inner products are accumulated using $\mu$ mantissa bits, and 0.9\% of adaptively selected inner products are recomputed in FP32 (corresponding to a threshold $\tau = 0.1$ in Sections~\ref{sec:theory}-\ref{sec:theory_llm}).
Notably, BF16 accumulation ($\mu = 7$) with adaptive recomputation deviates from uniform FP32 accumulation just as much as uniform TF32 accumulation ($\mu = 10$).
Meanwhile, the effective number of mantissa bits used per KQ inner product is about $8.9$ for the former against $10$ for the latter.\footnote{$1 \cdot 7 + 0.083 \cdot 23 = 8.909$}
In addition, Figure~\ref{fig:bits_lamp_vs_fake} shows that our method does indeed select critical-for-performance inner products, since the same number of random recomputations has no effect.
Furthermore, the ``look-ahead'' nature of our approach is vital: a non-look-ahead strategy that recomputes the most \emph{intrinsically} ill-conditioned inner products (those with a large ratio $|\Matrix{x}|^\trans |\Matrix{y}| / |\Matrix{x}^\trans \Matrix{y}|$) fails to deliver gains under the same recomputation budget.
Find a detailed description of our experiments and more numerical results in Section~\ref{sec:experiments} and Appendix~\ref{appendix:experiments}.

\textbf{Not sparse attention:}
Unlike sparse attention \cite{beltagy2020longformer, kitaev2020reformer}, which achieves efficiency by discarding low-scoring tokens without regard for the resulting probability mass shift, LAMP computes these scores in low precision.
By dynamically routing high FP precision only to numerically sensitive tokens, LAMP strictly bounds the worst-case rounding error of the softmax distribution.

\textbf{Benchmark:}
In this paper, we apply the LAMP framework to the \emph{offline} computation of softmax, and therefore the analysis and the resulting algorithm do not directly extend to the \emph{online}-softmax mechanism of FlashAttention \cite{dao2022flashattention} \footnote{Though the LAMP framework itself is not fundamentally incompatible with FlashAttention; see a follow-up work \cite{budzinskiy2026lampattention}}.
Consequently, our proposed algorithm, designed strictly based on the solution of the optimization problem \eqref{eq:lamp_problem} for the softmax function, falls short in long-context scenarios, because the entire vector of softmax probabilities needs to be materialized in memory.
Hence, we treat the ``strict'' LAMP algorithm for softmax as a benchmark that achieves an optimal balance between the recomputation rate and the accuracy gain.

\textbf{Practical LAMP attention:}
To compete with the state of the art, the ``strict'' LAMP needs to be relaxed in order to become compatible with the online softmax.
While such modification is a topic of our future work, we propose here a crucial intermediate step in this direction in the form of \emph{relaxed relative-threshold} LAMP (Subsection~\ref{subsec:rel_thresh}).
Our preliminary experiments indicate a marginal loss in performance compared to the ``strict'' LAMP.

\textbf{Hardware reliance:}
The numerical experiments carried out in this article simulate low-precision accumulation to allow for high granularity of mantissa bit-widths.
In a practical setting, it would be beneficial to pack multiple low-precision KQ inner products in a single 32-bit register (e.g., four 8-bit accumulators) and thereby increase the throughput.
As far as we know, current accelerators do not support native matrix-product accumulation in such packed formats. 
Therefore, our LAMP framework provides a mathematically grounded blueprint for next-generation AI accelerators that natively support packed low-precision accumulation.
For the same reason, we do not present runtime comparisons between our method and the state-of-the-art kernels tailored to modern hardware.

\textbf{Complementary idea to quantization:}
Let us stress that quantization is \emph{not} addressed in our work.
The proposed approach does not modify the operands of matrix products and does not reduce the memory footprint of a transformer \cite{frantar2023optq}---we aim to control the accuracy of \emph{computations} instead.
Consequently, our method should be viewed as complementary to quantization techniques.

\textbf{Extension to other architectures:}
The proposed LAMP inference strategy is derived from a theoretical analysis of the core building blocks of transformers.
While the underlying idea applies to arbitrary DNNs, the feasibility of its implementation is architecture-dependent.
Extending our approach to other DNN architectures would require a similar theoretical analysis of their constituent operations.
Given the strong empirical results for transformers, we view this as a natural direction for future work (including the simultaneous LAMP evaluation of all transformer nonlinearities).

\textbf{Extension to training:}
While the focus of this article is on inference, the same LAMP principle can be applied to compositions taking place during the backward pass.
This is a topic for future work.

\textbf{Validation on larger models:}
Our experiments focus on the family of GPT-2 transformer models.
While these models are moderate in size by current standards, they already exhibit the characteristic depth and compositional structure of modern transformers.
Importantly, the numerical effects targeted by our method are present at this scale, suggesting that they are likely to become more pronounced in larger architectures where softmax probabilities are expected to be more concentrated (a comparison of GPT-2 XL and GPT-2 small models in Appendix~\ref{subsec:pareto_models} supports this claim).
Consequently, this paper establishes a theoretical foundation and a behavioral proof of concept for LAMP.


\section{Floating-point evaluation of compositions}
\label{sec:theory}
Let $m,n,k \in \N$ and consider functions: $g : \Real^{k} \to \Real^{n}$ and $f : \Real^{n} \to \Real^{m}$.
Given $\Matrix{x} \in \Real^{k}$, our goal is to evaluate the composition $f(g(\Matrix{x}))$ in FP arithmetic.

\subsection{Baseline evaluation}
Let $\Matrix{y} = g(\Matrix{x}) \in \Real^{n}$ be the exact output of the inner function, and let $\hat{\Matrix{y}} \in \Real^{n}$ be its computed value.
Assuming that $g$ is evaluated with a mixed forward-backward stable algorithm \cite{higham2002accuracy}, rounding error analysis guarantees that the relative error is bounded by
\begin{equation}
\label{eq:g_baseline}
    \frac{|\hat{\Matrix{y}} - \Matrix{y}|}{|\Matrix{y}|} \leq \Matrix{c}_g \uh_{g} + \Oh{\uh_g^2}, \quad \Matrix{c}_g \in \Real_{+}^{n},
\end{equation}
where the absolute value, division, and comparison of vectors are componentwise, and $\uh_g$ denotes the unit round-off.
The nonnegative vector $\Matrix{c}_g$ describes how round-off errors are magnified during the computation of each component of $\Matrix{y}$ and depends on the rounding mode (e.g., deterministic or stochastic) and the specific algorithm used to evaluate $g$.

Next, $\hat{\Matrix{y}}$ is used as input for $f$, which is evaluated in precision $\uh_{f}$.
Denoting by $\hat{\Matrix{z}} \in \Real^{m}$ the computed value of $f(\hat{\Matrix{y}})$, we get by the same logic as in \eqref{eq:g_baseline} that
\begin{equation*}
    \frac{|\hat{\Matrix{z}} - f(\hat{\Matrix{y}})|}{|f(\hat{\Matrix{y}})|} \leq \Matrix{c}_f \uh_{f} + \Oh{\uh_f^2}, \quad \Matrix{c}_f \in \Real_{+}^{m}.
\end{equation*}
However, we need to compare $\hat{\Matrix{z}}$ with the exact $\Matrix{z} = f(\Matrix{y})$.
By triangle inequality,
\begin{equation*}
    \frac{|\hat{\Matrix{z}} - \Matrix{z}|}{|\Matrix{z}|}
    \leq \frac{|\hat{\Matrix{z}} - f(\hat{\Matrix{y}})| + |f(\hat{\Matrix{y}}) - \Matrix{z}|}{|f(\hat{\Matrix{y}})|} \Big(1 + \Oh{\uh_{f}}\Big).
\end{equation*}
We bound the first term by $\Matrix{c}_f \uh_{f}$ to first order.
To bound the second term, assume that $f$ is sufficiently regular in the neighborhood of $\hat{\Matrix{y}}$ and use Taylor's theorem to get a bound in terms of the Jacobian:
\begin{equation*}
    \frac{|f(\hat{\Matrix{y}}) - f(\Matrix{y})|}{|f(\hat{\Matrix{y}})|}
    \leq \frac{|\Matrix{J}_{f}(\hat{\Matrix{y}}) \diag{\hat{\Matrix{y}}}| \Matrix{c}_g}{|f(\hat{\Matrix{y}})|} \uh_{g} + \Oh{\uh_{g}^2},
\end{equation*}
where $\Matrix{J}_{f}(\hat{\Matrix{y}}) \in \Real^{m \times n}$ and $\diag{\hat{\Matrix{y}}} \in \Real^{n \times n}$ is diagonal.
As a result, we get
\begin{equation}
\label{eq:fg_baseline}
    \frac{|\hat{\Matrix{z}} - \Matrix{z}|}{|\Matrix{z}|}
    \leq \Matrix{c}_f \uh_{f} + \frac{|\Matrix{J}_{f}(\hat{\Matrix{y}}) \diag{\hat{\Matrix{y}}}| \Matrix{c}_g}{|f(\hat{\Matrix{y}})|} \uh_{g} + \Oh{\uh_{f}^2 + \uh_{g}^2}.
\end{equation}

\subsection{Refined evaluation}
\label{subsec:refined_eval}
The bound \eqref{eq:fg_baseline} follows from \eqref{eq:g_baseline} and therefore depends on $\Matrix{c}_g$ (i.e., the evaluation algorithm of $g$) and the assumption that every component of $g$ is computed in precision $\uh_g$.
Consider a more flexible setting where we may want to compute some of the components more accurately---with a more accurate algorithm or in higher precision---to ensure that the second term in \eqref{eq:fg_baseline} is not too large.
Let the nonzeros of $\Matrix{q} \in \{ 0,1 \}^{n}$ encode these components.

\subsubsection{More accurate algorithm}
We shall say that an evaluation algorithm is more accurate than the baseline if it leads to smaller entries in $\Matrix{c}_g$.
Let us denote by $0 \leq \epsilon \leq 1$ the corresponding gain factor. 
Then the bound \eqref{eq:g_baseline} becomes
\begin{equation*}
    \frac{|\hat{\Matrix{y}} - \Matrix{y}|}{|\Matrix{y}|} \leq \big(\Matrix{I} - \diag{\Matrix{q}}\big) \Matrix{c}_g \uh_{g} + \Oh{\uh_g^2 + \epsilon\uh_g},
\end{equation*}
leading to a modification of the second term in \eqref{eq:fg_baseline}:
\begin{equation*}
    \frac{|\Matrix{J}_{f}(\hat{\Matrix{y}}) \diag{\hat{\Matrix{y}}}| \big(\Matrix{I} - \diag{\Matrix{q}}\big) \Matrix{c}_g}{|f(\hat{\Matrix{y}})|} \uh_{g} + \Oh{\uh_{g}^2 + \epsilon\uh_g}.
\end{equation*}

For example, let $g(\Matrix{x}) = \Matrix{Ax}$ with $\Matrix{A} \in \Real^{n \times k}$ and $\Matrix{x} \in \Real^{k}$ be stored in precision $\uh_g$.
Evaluating $g$ in precision $\uh_g$, the basic multiplication algorithm has $\Matrix{c}_g = k \frac{|\Matrix{A}| |\Matrix{x}|}{|\Matrix{Ax}|}$ for deterministic rounding \cite{higham2002accuracy} and $\Matrix{c}_g \lesssim \sqrt{k} \frac{|\Matrix{A}| |\Matrix{x}|}{|\Matrix{Ax}|}$ with high probability for stochastic rounding \cite{connolly2021stochastic}.
Mixed-precision algorithms based on fused multiply-add can achieve $\Matrix{c}_g = \frac{|\Matrix{A}| |\Matrix{x}|}{|\Matrix{Ax}|}$ \cite{blanchard2020mixed}, and similarly for compensated-summation algorithms.
Therefore, the gain factor $\epsilon$ can be made small for large $k$.

\subsubsection{Higher precision}
As an alternative to a more accurate algorithm, we can use FP precision $\uh_g^2$ to compute and store the selected components.
Then \eqref{eq:g_baseline} turns into
\begin{equation*}
    \frac{|\hat{\Matrix{y}} - \Matrix{y}|}{|\Matrix{y}|} \leq \big(\Matrix{I} - \diag{\Matrix{q}}\big) \Matrix{c}_g \uh_{g} + \Oh{\uh_g^2}    
\end{equation*}
and leads to a similar modification in \eqref{eq:fg_baseline}.
This requires the evaluation algorithm of $f$ to process mixed-precision inputs.
Notably, mixed-precision matrix multiplication \cite{blanchard2020mixed} achieves $\Matrix{c}_g = \Matrix{0}$ when the output is stored in precision $\uh_{g}^2$, since no extra rounding is done at the end.

\subsection{Look-ahead mixed-precision evaluation}
\label{subsec:lamp_evaluation}
The two refinements in Subsection~\ref{subsec:refined_eval} lead to very similar componentwise rounding error bounds for the composition.
The accuracy of the inner computation manifests itself in the second term, which we aim to reduce as follows.
Denote
\begin{equation*}
    \Matrix{K}(f,\hat{\Matrix{y}}) = \Matrix{J}_{f}(\hat{\Matrix{y}}) \diag{\hat{\Matrix{y}}}, \quad \Matrix{M}(f,\hat{\Matrix{y}}) = \diag{f(\hat{\Matrix{y}})}^{-1} \Matrix{K}(f,\hat{\Matrix{y}}),
\end{equation*}
then we seek a binary vector $\Matrix{q} \in \{ 0,1 \}^{n}$ such that
\begin{equation}
\label{eq:lamp_comp}
    \kappa_{c}(f, \hat{\Matrix{y}}; \Matrix{q}) = \Norm[\big]{\Matrix{M}(f,\hat{\Matrix{y}}) (\Matrix{I} - \diag{\Matrix{q}})}{\infty,\infty} \leq \tau
\end{equation}
for a given threshold $\tau \geq 0$.
While this objective targets the componentwise error for the composition, a similar derivation for an $\ell_p$-based combination of input-output error metrics leads to
\begin{equation}
\label{eq:lamp_norm}
    \kappa_{p}(f, \hat{\Matrix{y}}; \Matrix{q}) = \frac{\Norm{\Matrix{K}(f,\hat{\Matrix{y}}) (\Matrix{I} - \diag{\Matrix{q}}) }{p,p}}{\Norm{f(\hat{\Matrix{y}})}{p}} \leq \tau.
\end{equation}
Let us discuss these objectives:
\begin{itemize}
    \item Objective \eqref{eq:lamp_comp} improves the bound \eqref{eq:fg_baseline}.
    Specifically, the second term in \eqref{eq:fg_baseline} can be bounded by $\One \tau \Norm{\Matrix{c}_g}{\infty} \uh_{g}$, and \eqref{eq:lamp_norm} has a similar effect on the corresponding rounding error bound.
    \item Both \eqref{eq:lamp_comp} and \eqref{eq:lamp_norm} can always be attained with $\Matrix{q} = \One$, i.e., when every component of the inner function $g$ is computed more accurately.
    \item When $\Matrix{q} = \Matrix{0}$, the quantity $\kappa_{c}(f, \hat{\Matrix{y}}; \Matrix{q})$ is the componentwise condition number of $f$\cite{gohberg1993mixed}.
    \item The exact $f(\hat{\Matrix{y}})$ and $\Matrix{J}_{f}(\hat{\Matrix{y}})$ are not known in practice, so their computed values will be used.
\end{itemize}

If the baseline $\hat{\Matrix{y}}$ with $\Matrix{q} = \Matrix{0}$ satisfies \eqref{eq:lamp_comp} or \eqref{eq:lamp_norm}, we deem the computation complete.
Otherwise, we try a different $\Matrix{q}$, which entails the recomputation of $\hat{\Matrix{y}}$.
Because such recomputations can be costly, we shall assume that the Jacobian is stable\footnote{If the Jacobian changes rapidly, the Hessian would need to be included in the analysis.} with respect to small variations in $\hat{\Matrix{y}}$ and fix
\begin{equation*}
    \kappa \in \{ \Matrix{q} \mapsto \kappa_{c}(f, \hat{\Matrix{y}}; \Matrix{q}),~\Matrix{q} \mapsto \kappa_{p}(f, \hat{\Matrix{y}}; \Matrix{q}) \}
\end{equation*}
corresponding to the baseline $\hat{\Matrix{y}}$.
Having found a suitable $\Matrix{q}$ that satisfies the bound $\kappa(\Matrix{q}) \leq \tau$, we will recompute more accurately those components of $\hat{\Matrix{y}}$ that are indexed by the nonzeros of $\Matrix{q}$.
To minimize the number of recomputations, we require $\Matrix{q}$ to be sparse.
This leads to our \emph{LAMP problem}:
\begin{equation}
\label{eq:lamp_problem}
    \Norm{\Matrix{q}}{0} \to \min \quad \text{s.t.} \quad \kappa(\Matrix{q}) \leq \tau.
\end{equation}
LAMP evaluation of a composition is summarized in Algorithm \ref{alg:lamp}.
We shall call the LAMP problem \eqref{eq:lamp_problem} \emph{componentwise} or \emph{$\ell_p$-based} depending on the underlying objective.

\begin{algorithm}[h]
\caption{LAMP evaluation of a composition}
\label{alg:lamp}
\begin{algorithmic}
    \STATE {\bfseries Input:} functions $f$ and $g$, variable $\Matrix{x}$, threshold $\tau \geq 0$, componentwise or normwise objective
    \STATE {\bfseries Output:} adaptively computed value $\hat{\Matrix{y}}$ of $g(\Matrix{x})$
    \STATE Compute $\hat{\Matrix{y}} \approx g(\Matrix{x})$ in FP arithmetic.
    \STATE Set up the function $\kappa$ in FP arithmetic.
    \STATE Find a solution $\Matrix{q}$ of the LAMP problem \eqref{eq:lamp_problem}.
    \STATE Recompute components of $\hat{\Matrix{y}}$ indexed by nonzeros of $\Matrix{q}$ more accurately.
\end{algorithmic}
\end{algorithm}


\section{Application to transformer inference}
\label{sec:theory_llm}
Algorithm~\ref{alg:lamp} relies on the possibility to compute individual components of $g$ separately.
Yet this is not always possible.
For instance, consider the softmax function
\begin{equation}
\label{eq:softmax}
    \mathrm{softmax}(\Matrix{x}) = \begin{bmatrix}
        \frac{\exp(x_1)}{\sum_{i = 1}^{k} \exp(x_i)} & \cdots & \frac{\exp(x_k)}{\sum_{i = 1}^{k} \exp(x_i)}
    \end{bmatrix}^\trans.
\end{equation}
If at least one component needs to be computed more accurately, every component has to be as well.
Meanwhile, the required property holds for matrix-vector products: it suffices to divide the matrix into two blocks and use different multiplication algorithms for each. 
When a bias term is present, it can be added to the accumulator in the required precision as well.

Matrix products are used in all DNN architectures.
We thus apply LAMP to the following question:
\begin{center}
    \emph{How to adapt matrix multiplication to the ensuing nonlinearity during DNN inference?}
\end{center}

What is the ``ensuing nonlinearity'' in question?
To target the accuracy of the end result of inference, an obvious choice is to set $f$ to be the remaining tail of the DNN.
However, the computation of the associated $\kappa(\Matrix{q})$ and the solution of the LAMP problem \eqref{eq:lamp_problem} are highly demanding.

We focus on the \emph{elementary transformer nonlinearities} (activation functions, layer normalization, softmax) and show that their LAMP problems \eqref{eq:lamp_problem} admit \emph{closed-form solutions}.

\subsection{Activation functions}
Consider an activation function $\varphi : \Real \to \Real$.
In DNNs, activation functions are applied entrywise:
\begin{equation*}
    f : \Real^n \to \Real^n, \quad f(\Matrix{y}) = \begin{bmatrix}
        \varphi(y_1) & \cdots & \varphi(y_n)
    \end{bmatrix}^\trans.
\end{equation*}
As a consequence, the matrix $\Matrix{M}$ is diagonal,
\begin{equation*}
    \Matrix{M}(f, \Matrix{y}) = \diag[\Big]{\tfrac{\varphi'(y_1)}{\varphi(y_1)}y_1, \ldots, \tfrac{\varphi'(y_n)}{\varphi(y_n)}y_n},
\end{equation*}
and the solution $\Matrix{q}$ of the componentwise LAMP problem \eqref{eq:lamp_problem} can be written in closed form: an entry of $\Matrix{q}$ is nonzero if and only if the corresponding diagonal entry of $\Matrix{M}$ exceeds $\tau$ in absolute value.
The diagonal structure of the Jacobian makes it possible to solve \eqref{eq:lamp_problem} immediately.

For activation functions, LAMP evaluation essentially repeats the mixed-precision accumulation of \cite{el2026mixed}.
The novelty of LAMP is generality: it applies to arbitrary compositions, and we prove that transformer-specific compositions are particularly suitable for LAMP evaluation.

\subsection{Layer normalization}
\label{subsec:layer_norm}
Layer normalization aims to stabilize the training of DNNs \cite{ba2016layer} by ``standardizing'' its input via\footnote{We focus on layer normalization without scale and bias, as they are an affine function applied after the nonlinearity.}
\begin{equation*}
    f : \Real^n \to \Real^n, \quad f(\Matrix{y}) = \sqrt{n} \frac{\Matrix{y} - \tfrac{1}{n} \One\One^\trans \Matrix{y}}{\Norm{\Matrix{y} - \tfrac{1}{n} \One\One^\trans \Matrix{y}}{2}}.
\end{equation*}
Its placement within a transformer can vary across specific architectures \cite{xiong2020layer}, the two conventional choices being \emph{post}normalization \cite{vaswani2017attention, devlin2019bert} and \emph{pre}normalization \cite{brown2020language, touvron2023llama, chowdhery2023palm}.
Yet neither places layer normalization right after a matrix product.

Another possible placement is between a (feedforward or attention) sublayer and a skip connection \cite{liu2022swin, olmo20252}, then layer normalization directly follows a matrix product and we can apply Algorithm~\ref{alg:lamp}.

Note that the shifting step in layer normalization is itself a matrix-vector product, and we can ``attach'' this matrix to the preceding matrix multiplication.
The remaining function
\begin{equation}
\label{eq:rms_norm}
    f : \Real^n \to \Real^n, \quad f(\Matrix{y}) = \sqrt{n} \frac{\Matrix{y}}{\Norm{\Matrix{y}}{2}}
\end{equation}
is \emph{root mean square} (RMS) layer normalization \cite{zhang2019root} and is often used on its own \cite{touvron2023llama, olmo20252}.
We analyze its componentwise LAMP problem \eqref{eq:lamp_problem}, which aims to control the relative error incurred in each individual entry of $f(\Matrix{y})$.

\begin{proposition}
\label{proposition:rms_norm}
Let $\Matrix{q} \in \{0,1\}^n$ be such that $\Matrix{q} \neq \One$, and denote by $\Omega$ its support.
For RMS layer normalization \eqref{eq:rms_norm},
\begin{equation*}
   \kappa_c(f, \Matrix{y}; \Matrix{q}) = 
   \begin{cases}
       2 \left( 1 - \frac{\min_{j \not\in \Omega} y_j^2}{\Norm{\Matrix{y}}{2}^2} \right) - \frac{\sum_{i \in \Omega} y_i^2}{\Norm{\Matrix{y}}{2}^2}, & |\Omega| \leq n-2,\\
       \max \left\{ \frac{y_j^2}{\Norm{\Matrix{y}}{2}^2}, 1 - \frac{y_j^2}{\Norm{\Matrix{y}}{2}^2} \right\}, & \{ 1, \ldots, n \} \setminus \Omega = \{ j \}.
   \end{cases}
\end{equation*}
\end{proposition}
\vspace{-0.5cm}
\begin{proof}
See Appendix~\ref{appendix:proposition:rms_norm}.
\end{proof}

Even though Proposition~\ref{proposition:rms_norm} provides exact values of $\kappa_c$ for each $\Matrix{q}$, finding the sparsest $\Matrix{q}$ to satisfy \eqref{eq:lamp_comp} is expensive for large $n$.
The following proposition shows that there exist simple solutions of \eqref{eq:lamp_problem}.

\begin{proposition}
\label{proposition:rms_norm_greedy}
Suppose that the entries of $\Matrix{y}$ are arranged as $y_1^2 \geq \cdots \geq y_n^2$.
Let $\Matrix{q} \in \{0,1\}^n$ be a solution of the componentwise LAMP problem \eqref{eq:lamp_problem} for RMS layer normalization \eqref{eq:rms_norm}.
If $\Norm{\Matrix{q}}{0} \leq n-3$ then a vector $\Matrix{q}' \in \{0,1\}^n$ given by 
\begin{equation*}
    \Matrix{q}' = \begin{bmatrix}
        1 & \cdots & 1 & 0 & \cdots & 0
    \end{bmatrix}^\trans, \quad \Norm{\Matrix{q}'}{0} = \Norm{\Matrix{q}}{0} + q_n,
\end{equation*}
also satisfies $\kappa_c(f, \Matrix{y}; \Matrix{q}') \leq \tau$.
\end{proposition}
\vspace{-0.3cm}
\begin{proof}
See Appendix~\ref{appendix:proposition:rms_norm_greedy}.
\end{proof}

Proposition~\ref{proposition:rms_norm_greedy} guarantees that an almost-the-sparsest solution of the componentwise LAMP problem \eqref{eq:lamp_problem} has a closed-form expression and can be computed greedily: we need to sort the entries of $\Matrix{y}$ in descending order according to their squares, pick the smallest $s$ such that
\begin{equation*}
    \sum_{i = 1}^{s} y_i^2 + 2 y_n^2 \geq (2 - \tau) \Norm{\Matrix{y}}{2}^2,
\end{equation*}
and form $\Matrix{q}$ based on the initial positions of the indices.
For a spread-out vector with $y_1^2 = \cdots = y_{n-1}^2 = 1$ and $y_n = 0$, we get $s = \lceil (2 - \tau) (n-1) \rceil$.
On the contrary, $s = 1$ when $y_1^2 = 1$ and $y_2 = \cdots = y_n = 0$.
Thus, vectors with massive outliers require a small number of recomputations.

\subsection{Attention and softmax}
The self-attention mechanism is a distinctive feature of transformers \cite{vaswani2017attention}.
In the simplest case, attention is computed as $\Matrix{V}\cdot\mathrm{softmax}(\Matrix{K}^\trans \Matrix{Q})$. This is a composition of three functions: the product of the key and query matrices, the softmax function \eqref{eq:softmax} applied columnwise, and the product of the value matrix and the output of softmax.

When the value-softmax product is followed by layer normalization, its LAMP evaluation can be performed as in Subsection~\ref{subsec:layer_norm}.
Here, we study the LAMP problem \eqref{eq:lamp_problem} for
\begin{equation*}
    f : \Real^n \to \Real^n, \quad f(\Matrix{y}) = \mathrm{softmax}(\Matrix{y}).
\end{equation*}

There is a remarkable distinction between the componentwise LAMP problems for RMS layer normalization and softmax.
Whereas the former admits a simple solution at the cost of one additional index, Appendix~\ref{appendix:softmax_counterexamples} shows that enlarging an optimal-sized greedy index set by any fixed or percentage-based margin is insufficient to ensure the componentwise LAMP objective \eqref{eq:lamp_comp} for softmax.

Because highly negative KQ inner products create pathological failure cases for the componentwise LAMP, we pivot to the $\ell_1$-based objective \eqref{eq:lamp_norm}, which naturally bounds the total probability mass shift.
Recall the notation $\Matrix{z} = f(\Matrix{y)}$.

\begin{proposition}
\label{proposition:softmax}
Let $\Matrix{q} \in \{0,1\}^n$ be such that $\Matrix{q} \neq \One$, and denote by $\Omega$ its support.
For softmax \eqref{eq:softmax},
\begin{equation*}
    \kappa_1(f, \Matrix{y}; \Matrix{q}) = 2 \max_{j \not\in \Omega} z_j (1 - z_j) |y_j|.
\end{equation*}
\end{proposition}
\vspace{-0.5cm}
\begin{proof}
See Appendix~\ref{appendix:proposition:softmax}.
\end{proof}

The optimal solution of the LAMP objective is simple: the $j$th entry of $\Matrix{q}$ is nonzero if and only if
\begin{equation}
\label{eq:strict_lamp}
    2 z_j (1 - z_j) |y_j| > \tau.
\end{equation}
This condition postulates that the tiniest probabilities are very stable (unlike in the case of componentwise LAMP).
For an extremely concentrated distribution where $\Matrix{z}$ is close to a standard basis vector, no recomputations are needed.
When an attention head is ``confused'' and produces multiple equally probable outcomes, these can necessitate more accurately computed KQ inner products.

As noted in Subsection~\ref{sec:contributions}, computing the optimal solution \eqref{eq:strict_lamp} requires a fully materialized vector of softmax probabilities.
In Subsection~\ref{subsec:rel_thresh}, we bring the ``strict'' LAMP closer to the one-pass softmax.


\section{Numerical experiments}
\label{sec:experiments}

The code used for the experiments is publicly available.\footnote{https://github.com/sbudzinskiy/LAMP-LLM}
The computations were carried out on a cloud-based GPU equipped with 80 GB of VRAM and required approximately 60 GPU-hours.

\subsection{Custom floating-point format}
To focus on precision and put aside the issue of overflows, we simulate low-precision matrix products using a custom \emph{partial single} FP format.
For every $\mu \in \{ 1, \ldots, 23 \}$, we define a format $\PS(\mu)$ with $\mu$ mantissa bits, 8 exponent bits, and one sign bit.
This format is equivalent to FP32 when $\mu = 23$, to TF32 when $\mu = 10$, and to BF16 when $\mu = 7$.
In code, we implement $\PS(\mu)$ numbers via FP32 numbers rounded to $\mu$ mantissa bits according to the round-to-nearest-ties-to-even mode.

To multiply matrices with input-output formats $\PS(\mu_A) \times \PS(\mu_B) \to \PS(\mu_C)$, we accumulate inner products as $\mathrm{round}(c + a \cdot b)$, where the scalar multiplication and addition are in FP32.

\subsection{Experimental setting}
We validate the LAMP evaluation of attention according to \eqref{eq:strict_lamp} in experiments with the GPT-2 XL model and the OpenWebText\footnote{https://huggingface.co/datasets/Skylion007/openwebtext} dataset.
Find more experiments in Appendix~\ref{appendix:experiments}.

To assess the impact of LAMP evaluation on inference, we compute the mean Kullback--Leibler (KL) divergence between the probability distributions output by a reference model and a test model over 100 sequences of 1024 tokens each.
We also look at the flip rate, i.e., how often the most probable predictions of the reference and test models differ.

Our reference model uses FP32 inference uniformly for all FP operations.
The test models perform the KQ products in $\PS(\mu)$ and recompute those selected by the LAMP solution \eqref{eq:strict_lamp} in FP32.
We keep track of how many inner products are recomputed; to get the recomputation rate, we divide by the number of KQ inner products in the ``causal mask.''

\subsection{Numerical results: Proof of concept}

\begin{figure}[t]
\centering
\begin{subfigure}[b]{\linewidth}
\centering
	\includegraphics[width=\linewidth]{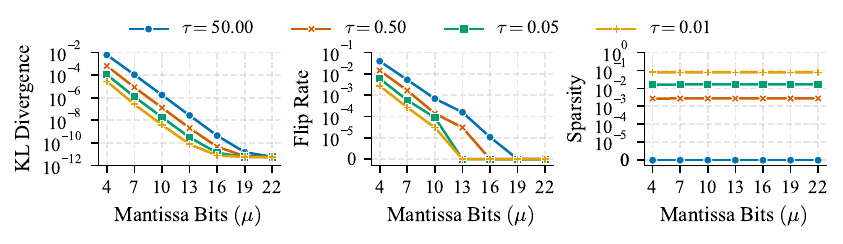}
\end{subfigure}\vspace{-0.3cm}\\
\begin{subfigure}[b]{\linewidth}
\centering
	\includegraphics[width=\linewidth]{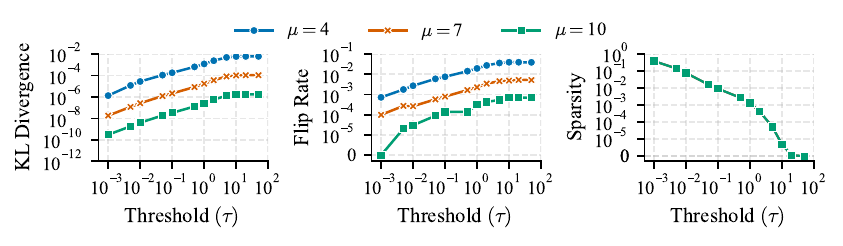}
\end{subfigure}
    \caption{Performance of mixed-precision GPT-2 XL inference on the OpenWebText dataset with LAMP evaluation of KQ inner products.}
\label{fig:bits_tau_1x3}
\end{figure}

The plots in Figure~\ref{fig:bits_tau_1x3} demonstrate that our method works as intended: as the threshold $\tau$ decreases, it improves the accuracy of inference and increases the number of recomputations required.
Our method achieves consistent $12\times$, $83\times$, and $385\times$ reductions in KL divergence at recomputation rates of only 0.3\%, 1.6\%, and 7.6\%, respectively.
Similar improvements hold for the flip rate, and we observe an exponential decay of the two metrics.
An interesting observation is that the recomputation rate hardly depends on $\mu$, which serves as circumstantial evidence for the stability of the Jacobian to small input variations (Subsection~\ref{subsec:lamp_evaluation}).
We repeat that the \emph{choice} of the components to be recomputed has a crucial impact on accuracy gains (see Figure~\ref{fig:bits_lamp_vs_fake} and Appendix~\ref{subsec:fake_lamp}).

These results are a qualitative proof of concept for the LAMP evaluation of softmax in attention, as it reaches high inference accuracies relative to the ground truth with the majority of KQ inner products performed in low precision.
The additional experiments in Appendix~\ref{appendix:experiments} strengthen this conclusion.

\subsection{Relaxed relative-threshold LAMP for softmax}
\label{subsec:rel_thresh}
Consider a relaxation of the optimal LAMP solution \eqref{eq:strict_lamp} for softmax, obtained by omitting $1 - z_j$.
One negligible effect incurred by this relaxation is that strictly dominant tokens with $z_j \approx 1$ can now be considered numerically sensitive, whereas they are perfectly stable in the ``strict'' formulation \eqref{eq:strict_lamp}.
A more significant effect emerges once we switch to a relative threshold $0 \leq \tau < 1$,
\begin{equation}
\label{eq:rrt_lamp}
    |y_j| e^{y_j} > \tau \max_{1 \leq i \leq n} |y_i| e^{y_i}.
\end{equation}
The defining advantage of the relaxed relative-threshold LAMP \eqref{eq:rrt_lamp} is its complete independence from the normalization constant $\sum_i e^{y_i}$.
This decoupling is a critical achievement for our future work on applying the LAMP framework to the one-pass design of FlashAttention.

\begin{figure}[h!]
    \centering
    \includegraphics[width=0.97\linewidth]{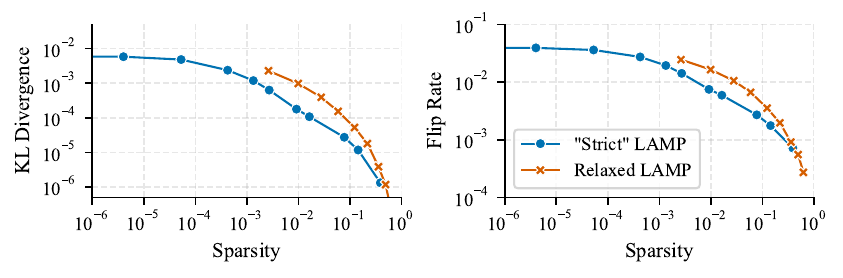}
    
    \caption{Comparison of Pareto boundaries of ``strict'' LAMP \eqref{eq:strict_lamp} and relaxed LAMP \eqref{eq:rrt_lamp} applied to GPT-2 XL with $\mu = 4$ mantissa bits for KQ accumulation and validated on the OpenWebText dataset.}
    \label{fig:pareto_relax}
\end{figure}

In Figure~\ref{fig:pareto_relax}, we compare the Pareto boundaries for the optimal and relaxed LAMP solutions, i.e., plotting the accuracy metrics (KL divergence and flip rate) against the efficiency metric (recomputation rate).
As the theoretically optimal solution, the ``strict'' LAMP \eqref{eq:strict_lamp} achieves a perfect balance between accuracy and efficiency, and therefore its Pareto boundaries are lower and serve as a benchmark.
Meanwhile, the Pareto boundaries of the relaxed LAMP \eqref{eq:rrt_lamp} exhibit only a marginal upward shift, which is an illustration of its almost-optimality.
These results validate the stability of our relaxation, paving the way for the efficient and effective integration of the LAMP framework into FlashAttention.

\section*{Author Contributions}
SB conceived the approach, formulated the research problem, and carried out the formal analysis, experimentation, and implementation. SB wrote the original draft of the manuscript.
MG  and TY contributed to the analysis and implementation and assisted with proofreading. YHT contributed to project administration and to reviewing and editing the manuscript.
YL, WF, and FW contributed to project administration. PP supervised the project as laboratory head and reviewed and edited the manuscript.

\section*{Acknowledgments}
This work was carried out in the framework of a research project funded by Huawei Technologies Ltd.
We are grateful to El-Mehdi El Arar, Silviu-Ioan Filip, Theo Mary, and Elisa Riccietti for fruitful discussions.


\bibliography{main}
\bibliographystyle{abbrv}


\newpage
\appendix
\section{Proofs}

\subsection{Proof of Proposition~\ref{proposition:rms_norm}}
\label{appendix:proposition:rms_norm}
Direct calculation yields
\begin{equation*}
    \Matrix{J}_f(\Matrix{y}) = \Matrix{I} - \frac{\Matrix{y} \Matrix{y}^\trans}{\Norm{\Matrix{y}}{2}^2}, \quad \Matrix{M}(f, \Matrix{y}) = \Matrix{I} - \frac{\One \Matrix{y}^\trans}{\Norm{\Matrix{y}}{2}^2} \diag{\Matrix{y}}.
\end{equation*}
Denote by $s_l$ the $l$th absolute row-sum of $\Matrix{M}(f, \Matrix{y}) \big(\Matrix{I} - \diag{\Matrix{q}}\big)$:
\begin{equation*}
    s_l = \begin{cases}
        \tfrac{1}{\Norm{\Matrix{y}}{2}^2} \sum_{j \not\in \Omega} y_j^2, & l \in \Omega,\\
        \tfrac{1}{\Norm{\Matrix{y}}{2}^2} \sum_{j \not\in \Omega \cup \{l\}} y_j^2 + 1 - \tfrac{y_l^2}{\Norm{\Matrix{y}}{2}^2}, & l \not\in \Omega.
    \end{cases}
\end{equation*}
Rewriting $\sum_{j \not\in \Omega} y_j^2 = \Norm{\Matrix{y}}{2}^2 - \sum_{i \in \Omega} y_i^2$ and taking the maximum of $s_l$ over $l$, we get
\begin{equation*}
    \kappa_{c}(f, \Matrix{y}; \Matrix{q}) = \max\left\{ 2 \left( 1 - \frac{\min_{j \not\in \Omega} y_j^2}{\Norm{\Matrix{y}}{2}^2} \right), 1 \right\} - \frac{\sum_{i \in \Omega} y_i^2}{\Norm{\Matrix{y}}{2}^2}.
\end{equation*}
Note that $1 > 2 \Big( 1 - \frac{\min_{j \not\in \Omega} y_j^2}{\Norm{\Matrix{y}}{2}^2} \Big)$ if and only if $\min_{j \not\in \Omega} y_j^2 > \frac{1}{2} \Norm{\Matrix{y}}{2}^2$.
When $|\Omega| \leq n - 2$, this would imply that
\begin{equation*}
    \sum_{j \not\in \Omega} y_j^2 > \frac{1}{2}(n - |\Omega|) \Norm{\Matrix{y}}{2}^2 \geq \Norm{\Matrix{y}}{2}^2,    
\end{equation*}
which is impossible.
This contradiction proves the first formula.
The second formula follows from the general expression.

\subsection{Proof of Proposition~\ref{proposition:rms_norm_greedy}}
\label{appendix:proposition:rms_norm_greedy}
Denote by $\Omega$ and $\Omega'$ the supports of $\Matrix{q}$ and $\Matrix{q}'$, respectively.
By Proposition~\ref{proposition:rms_norm}, as $|\Omega'| \leq n-2$, we need to show that
\begin{equation*}
    2 \left( 1 - \frac{y_n^2}{\Norm{\Matrix{y}}{2}^2} \right) - \frac{\sum_{i \in \Omega'} y_i^2}{\Norm{\Matrix{y}}{2}^2} \leq \tau.
\end{equation*}
Since $|\Omega| < n-2$ and $\Matrix{q}$ satisfies the componentwise LAMP constraint, it suffices to show that
\begin{equation*}
    \sum_{i \in \Omega'} y_i^2 + 2y_n^2
    \geq \sum_{i \in \Omega} y_i^2 + 2\min_{j \not\in \Omega} y_j^2.
\end{equation*}
Note that $\Omega'$ is constructed in such a way that for every $\Tilde{\Omega} \subset \{1, \ldots, n\}$ with $|\Tilde{\Omega}| = |\Omega'|$, it holds that $\sum_{i \in \Omega'} y_i^2 \geq \sum_{i \in \Tilde{\Omega}} y_i^2$.
When $n \not\in \Omega$, we have $\min_{j \not\in \Omega} y_j^2 = y_n^2$, and the desired inequality follows trivially since $|\Omega| = |\Omega'|$.
When $n \in \Omega$, we have
\begin{equation*}
    \sum_{i \in \Omega} y_i^2 + 2\min_{j \not\in \Omega} y_j^2
    \leq  y_n^2 + \sum_{i \in \Omega \setminus \{ n \}} y_i^2 
        + \min_{\substack{j,l \not\in \Omega \\ j \neq l}} ( y_j^2 + y_{l}^2 )
    \leq y_n^2 + \sum_{i \in \Omega'} y_i^2
    \leq 2y_n^2 + \sum_{i \in \Omega'} y_i^2,
\end{equation*}
where in the second inequality we apply the majorizing property of $\Omega'$ to the set $\Tilde{\Omega} = \Omega \setminus \{ n \} \cup \{ j_\ast, l_\ast \}$, with minimizing indices $j_\ast$ and $l_\ast$, and note that $|\Tilde{\Omega}| = |\Omega'|$.

\subsection{Proof of Proposition~\ref{proposition:softmax}}
\label{appendix:proposition:softmax}
Direct calculation yields
\begin{equation*}
    \Matrix{J}_f(\Matrix{y}) = \diag{\Matrix{z}} - \Matrix{z} \Matrix{z}^\trans, \quad \Matrix{K}(f, \Matrix{y}) = \big( \diag{\Matrix{z}} - \Matrix{z} \Matrix{z}^\trans \big) \diag{\Matrix{y}}.
\end{equation*}
Then
\begin{align*}
    \kappa_1(f, \Matrix{y}; \Matrix{q}) &= \Norm{\Matrix{K}(f, \Matrix{y}) (\Matrix{I} - \diag{\Matrix{q}})}{1,1} = \max_{j \not\in \Omega} \sum_i |z_j y_j \delta_{i,j} - z_i z_j y_j| \\
    &= \max_{j \not\in \Omega} \Big[ z_j (1 - z_j) |y_j| + z_j |y_j| \sum_{i \neq j} z_i \Big] = \max_{j \not\in \Omega} \Big[ z_j (1 - z_j) |y_j| + z_j |y_j| (1 - z_j) \Big].
\end{align*}

\clearpage
\section{Componentwise LAMP for softmax: Counterexamples}
\label{appendix:softmax_counterexamples}
Let us derive an explicit expression for $\kappa_c(f, \Matrix{y}; \Matrix{q})$.
Let $\Omega$ be the support of $\Matrix{q}$, then
\begin{align*}
    \kappa_c(f, \Matrix{y}; \Matrix{q}) &= \Norm{\Matrix{M}(f,\Matrix{y}) (\Matrix{I} - \diag{\Matrix{q}})}{\infty, \infty} \\
    &= \max_{1 \leq i \leq n} \sum_{j \not\in \Omega} |\delta_{i,j} - z_j| |y_j| = \sum_{j \not\in \Omega} z_j |y_j| + \max_{i \not\in \Omega} (1 - 2 z_i) |y_i|.
\end{align*}
The formula naturally contains auxiliary vectors $\Matrix{u} \in \Real^n$ with $u_j = z_j |y_j|$ and $\Matrix{v} \in \Real^n$ with $v_i = (1 - 2 z_i) |y_i|$.
If either $\Matrix{u}$ or $\Matrix{v}$ were identically zero, the componentwise LAMP problem \eqref{eq:lamp_problem} would admit a simple exact solution based on thresholding or sorting, respectively.
This is not the case in general, though, and we prove that the exact solution $\Matrix{q}$ of \eqref{eq:lamp_problem} cannot be approximated with a greedy surrogate based on sorting the entries of $\Matrix{u}$ or picking the largest entries of $\Matrix{v}$.
Specifically, we design two families of examples that illustrate that for any $s \in \N$, taking a greedily constructed $\Matrix{q}'$ with support $\Omega'$ of cardinality $|\Omega'| = |\Omega| + s$ does not guarantee $\kappa_{c}(f, \Matrix{y}; \Matrix{q}') \leq \tau$.
Because the choice of $s$ is independent of $|\Omega|$, this simultaneously shows that increasing the cardinality by any arbitrary percentage is not a viable strategy either.

In the first example, the optimal solution is to select the most negative entries of $\Matrix{y}$ to minimize the second term $\max_{i \not\in \Omega} v_i$, whereas the greedy strategy based on the vector $\Matrix{u}$ ignores these entries because of their tiny probabilities.

\begin{proposition}
Let $n_0, s \in \N$ and set $n = 2 n_0 + s$.
There exists a vector $\Matrix{y} \in \Real^{n}$ and a threshold $\tau > 0$ such that
\begin{equation*}
    n_0 = \min \set{\Norm{\Matrix{q}}{0}}{\kappa_{c}(f, \Matrix{y}; \Matrix{q}) \leq \tau},
\end{equation*}
but the heuristic $\Matrix{q}' \in \{ 0,1 \}^n$ formed greedily by selecting the largest $n_0 + s$ entries of either $\Matrix{u}$ or $\Matrix{z}$ satisfies $\kappa_{c}(f, \Matrix{y}; \Matrix{q}') > \tau$.
\end{proposition}
\begin{proof}
Fix $\alpha \geq 3$ and define a vector $\Matrix{y} \in \Real^n$ with entries
\begin{equation*}
    y_i = \begin{cases}
        -\alpha, & 1 \leq i \leq n_0, \\
        -1, & n_0 < i \leq n.
    \end{cases}
\end{equation*}
Denote $Z = n_0 e^{-\alpha} + (n_0 + s) e^{-1}$.
Then the vectors $\Matrix{z}$, $\Matrix{u}$, and $\Matrix{v}$ evaluate to
\begin{equation*}
    z_i = \begin{cases}
        \frac{1}{Z e^{\alpha}}, & 1 \leq i \leq n_0, \\
        \frac{1}{Z e}, & n_0 < i \leq n,
    \end{cases} \quad
    u_i = \begin{cases}
        \frac{\alpha}{Z e^{\alpha}}, & 1 \leq i \leq n_0, \\
        \frac{1}{Z e}, & n_0 < i \leq n,
    \end{cases} \quad
    v_i = \begin{cases}
        \alpha (1 - \frac{2}{Z e^{\alpha}}), & 1 \leq i \leq n_0, \\
        1 - \frac{2}{Z e}, & n_0 < i \leq n.
    \end{cases}
\end{equation*}
Note that the function $x e^{-x}$ reaches its maximum at $x = 1$.
Define $\Matrix{q} \in \{0,1\}^n$ with support $\Omega = \{ 1, \ldots, n_0 \}$ and set
\begin{equation*}
    \tau = \kappa_{c}(f, \Matrix{y}; \Matrix{q}) = (n_0 + s) \frac{1}{Z e} + 1 - \frac{2}{Z e} < 1 + \frac{n_0 + s}{Z e} < 1 + \frac{n_0 + s}{(n_0 + s) e^{-1} e} = 2.
\end{equation*}
For any $\Matrix{\Tilde{q}} \in \{0,1\}^n$ with $\Norm{\Matrix{\Tilde{q}}}{0} < n_0$, the complement of $\tilde{\Omega}$ contains at least one $i \in \{ 1, \ldots, n_0 \}$, and thus
\begin{equation*}
    \kappa_{c}(f, \Matrix{y}; \Matrix{\tilde{q}}) > \alpha \Big(1 - \frac{2}{Z e^{\alpha}}\Big) \geq \alpha \Big( 1 - \frac{2}{1 + 2 e^{\alpha-1}} \Big) > 2 > \tau = \kappa_{c}(f, \Matrix{y}; \Matrix{q}),
\end{equation*}
proving that $\Matrix{q}$ is optimal for the chosen $\tau$.
By construction, the heuristic vector $\Matrix{q'}$ has support $\Omega' = \{ n_0 + 1, \ldots, n \}$, so
\begin{equation*}
    \kappa_{c}(f, \Matrix{y}; \Matrix{q}') = n_0 \frac{\alpha}{Z e^{\alpha}} + \alpha \Big(1 - \frac{2}{Z e^{\alpha}} \Big) > \alpha \Big(1 - \frac{2}{Z e^{\alpha}} \Big) > 2 > \tau. \qedhere
\end{equation*}
\end{proof}

In the second example, the optimal solution is to take the largest entries of $\Matrix{y}$ so as to minimize the first term $\sum_{j \not\in \Omega} u_j$, but the greedy strategy based on the vector $\Matrix{v}$ ignores these entries because their probabilities force $1 - 2 z_i$ to be small.

\begin{proposition}
Let $n_0, s \in \N$ and set $n = 2 n_0 + s$.
There exists a vector $\Matrix{y} \in \Real^{n}$ and a threshold $\tau > 0$ such that
\begin{equation*}
    n_0 = \min \set{\Norm{\Matrix{q}}{0}}{\kappa_{c}(f, \Matrix{y}; \Matrix{q}) \leq \tau},
\end{equation*}
but the heuristic $\Matrix{q}' \in \{ 0,1 \}^n$ formed greedily by selecting the largest $n_0 + s$ entries of $\Matrix{v}$ satisfies $\kappa_{c}(f, \Matrix{y}; \Matrix{q}') > \tau$.
\end{proposition}
\begin{proof}
Consider a constant
\begin{equation*}
    \alpha = \frac{(n_0 + s) (5 n_0 - 4)}{4s} \log\Big(\frac{n_0 + s}{n_0}\Big) > 0
\end{equation*}
and define a vector $\Matrix{y} \in \Real^n$ with entries
\begin{equation*}
    y_i = \begin{cases}
        \alpha + \log\big(\frac{n_0 + s}{n_0}\big), & 1 \leq i \leq n_0, \\
        \alpha, & n_0 < i \leq n.
    \end{cases}
\end{equation*}
Then
\begin{gather*}
    z_i = \begin{cases}
        \frac{1}{2 n_0}, & 1 \leq i \leq n_0, \\
        \frac{1}{2 (n_0 + s)}, & n_0 < i \leq n,
    \end{cases} \quad
    u_i = \begin{cases}
        \frac{1}{2 n_0} \Big( \alpha + \log\big(\frac{n_0 + s}{n_0}\big) \Big), & 1 \leq i \leq n_0, \\
        \frac{1}{2 (n_0 + s)} \alpha, & n_0 < i \leq n,
    \end{cases}\\
    v_i = \begin{cases}
        \big( 1 - \frac{1}{n_0} \big) \Big( \alpha + \log\big(\frac{n_0 + s}{n_0}\big) \Big), & 1 \leq i \leq n_0, \\
        \big( 1 - \frac{1}{n_0 + s} \big) \alpha, & n_0 < i \leq n.
    \end{cases}
\end{gather*}
Direct calculation yields $v_i > v_j$ for $n_0 < i \leq n$ and $1 \leq j \leq n_0$.
Let $\Matrix{q} \in \{0,1\}^n$ have support $\Omega = \{ 1, \ldots, n_0 \}$ and set
\begin{equation*}
    \tau = \kappa_{c}(f, \Matrix{y}; \Matrix{q}) = \frac{\alpha}{2} + \Big( 1 - \frac{1}{n_0 + s} \Big) \alpha.
\end{equation*}
For any $\Matrix{\Tilde{q}} \in \{0,1\}^n$ with $\Norm{\Matrix{\Tilde{q}}}{0} < n_0$, the complement of $\tilde{\Omega}$ contains at least one index from each of the two groups, so
\begin{equation*}
    \kappa_{c}(f, \Matrix{y}; \Matrix{\tilde{q}}) = \sum_{j \not\in \Tilde{\Omega}} u_j + \Big( 1 - \frac{1}{n_0 + s} \Big) \alpha \geq \frac{1}{2 n_0} \Big( \alpha + \log\Big(\frac{n_0 + s}{n_0}\Big) \Big) + \frac{n_0 + s}{2 (n_0 + s)} \alpha + \Big( 1 - \frac{1}{n_0 + s} \Big) \alpha > \tau,
\end{equation*}
proving that $\Matrix{q}$ is optimal for the chosen $\tau$.
By construction, the heuristic vector $\Matrix{q'}$ has support $\Omega' = \{ n_0 + 1, \ldots, n \}$, so
\begin{align*}
    \kappa_{c}(f, \Matrix{y}; \Matrix{q}') &= \frac{1}{2} \Big( \alpha + \log\Big(\frac{n_0 + s}{n_0}\Big) \Big) + \Big( 1 - \frac{1}{n_0} \Big) \Big( \alpha + \log\Big(\frac{n_0 + s}{n_0}\Big) \Big) - \tau + \tau \\
    &= \Big( \frac{3}{2} - \frac{1}{n_0} \Big) \log\Big(\frac{n_0 + s}{n_0}\Big) - \frac{s \alpha}{n_0 (n_0 + s)} + \tau \\
    &= \Big( \frac{3}{2} - \frac{1}{n_0} - \frac{5 n_0 - 4}{4 n_0} \Big) \log\Big(\frac{n_0 + s}{n_0}\Big) + \tau = \frac{1}{4} \log\Big(\frac{n_0 + s}{n_0}\Big) + \tau > \tau. \qedhere
\end{align*}
\end{proof}

\clearpage
\section{Additional numerical experiments}
\label{appendix:experiments}

\subsection{Comparison of performance: Different datasets}
\begin{figure}[ht!]
    \centering
    \includegraphics[width=\linewidth]{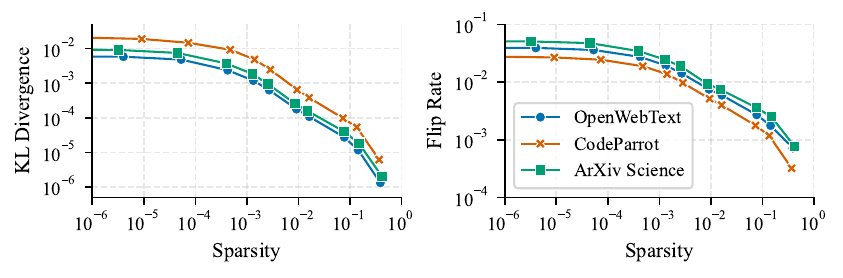}

    \caption{Comparison of Pareto boundaries of LAMP \eqref{eq:strict_lamp} applied to GPT-2 XL with $\mu = 4$ mantissa bits for KQ accumulation and validated on different datasets.}
    \label{fig:pareto_datasets}
\end{figure}

In the main text, we presented experiments with the GPT-2 XL model and the OpenWebText\footnote{https://huggingface.co/datasets/Skylion007/openwebtext} dataset.
Here, we add two more datasets for comparison: CodeParrot\footnote{https://huggingface.co/datasets/codeparrot/codeparrot-clean} and ArXiv.\footnote{https://huggingface.co/datasets/ccdv/arxiv-summarization}
The comparison of Pareto boundaries in Figure~\ref{fig:pareto_datasets} shows that the behavior of the proposed LAMP method \eqref{eq:strict_lamp} for softmax remains almost the same across different datasets, suggesting that our approach is ``input-agnostic.'' 

\subsection{Comparison of performance: Different models}
\label{subsec:pareto_models}
\begin{figure}[ht!]
    \centering
    \includegraphics[width=\linewidth]{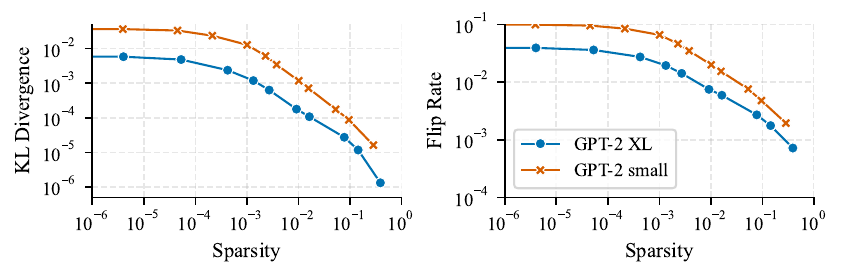}
    
    \caption{Comparison of Pareto boundaries of LAMP \eqref{eq:strict_lamp} applied to GPT-2 XL and GPT-2 small with $\mu = 4$ mantissa bits for KQ accumulation and validated on the OpenWebText dataset.}
    \label{fig:pareto_models}
\end{figure}

Next, we validate the proposed method on a different transformer model---the GPT-2 small model.
The numerical results in Figure~\ref{fig:pareto_models} show that LAMP evaluation of attention performs better (has a lower Pareto boundary) for the GPT-2 XL model than for the GPT-2 small model, supporting our vision that the performance of LAMP is likely to improve further for larger transformer architectures.

\subsection{Comparison of performance: Permuted sequences of tokens}
\begin{figure}[ht!]
    \centering
    \includegraphics[width=\linewidth]{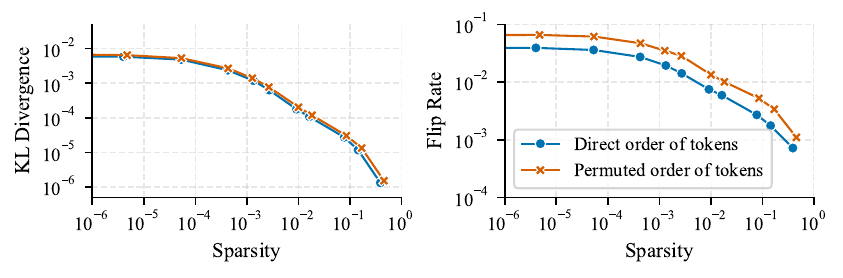}
    
    \caption{Comparison of Pareto boundaries of LAMP \eqref{eq:strict_lamp} applied to GPT-2 XL with $\mu = 4$ mantissa bits for KQ accumulation and validated on the OpenWebText dataset with direct and permuted tokens.}
    \label{fig:pareto_shuffled}
\end{figure}

To isolate the effect of word order, we construct three modified datasets by permuting the tokens in each sequence at random.
This eliminates all sequential dependencies while preserving the unigram distribution, i.e., the new sequences no longer correspond to a coherent text yet consist of the same tokens in a different order.
Figure~\ref{fig:pareto_shuffled} shows that LAMP inference performs equally well on such ``incoherent'' data, supporting our claim that the proposed method is ``input-agnostic.''
An interesting observation is that while the Pareto boundaries of the KL divergence essentially overlap, the Pareto boundary of the flip rate shifted upwards slightly for the permuted tokens.

\subsection{Comparison of performance: Random recomputation}
\label{subsec:fake_lamp}
\begin{figure}[ht!]
    \centering
    \includegraphics[width=\linewidth]{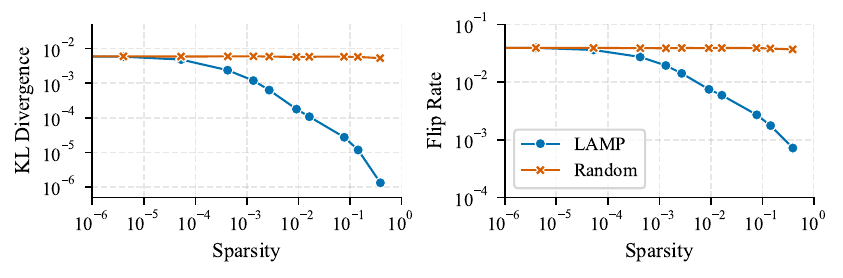}
    
    \caption{Comparison of Pareto boundaries of LAMP \eqref{eq:strict_lamp} and its analogue with randomly selected recomputations applied to GPT-2 XL with $\mu = 4$ mantissa bits for KQ accumulation and validated on the OpenWebText dataset.}
    \label{fig:pareto_fake}
\end{figure}

The proposed LAMP inference adaptively chooses those KQ inner products that need to be recomputed more accurately.
To show that it is not just the \emph{number} of recomputations that is significant, but the actual \emph{choice} of them, we perform experiments with \emph{random} recomputations: the number of inner products to redo is chosen by LAMP, but the specific inner products are selected at random.
An experiment of this sort was presented in Figure~\ref{fig:bits_lamp_vs_fake}.
The results in Figure~\ref{fig:pareto_fake} leave no doubt that the adaptive choice of the recomputations is the crux of our method, since random recomputations do not lead to any improvements.

\subsection{Perplexity and length-normalized thresholding}
Previously, we focused on the KL divergence and flip rate as the accuracy metrics to validate LAMP; here, we consider the perplexity attained on GSM8k Math,\footnote{https://huggingface.co/datasets/openai/gsm8k} WikiText-2,\footnote{https://huggingface.co/datasets/Salesforce/wikitext} and CodeParrot.\footnote{https://huggingface.co/datasets/codeparrot/codeparrot-clean}
For each dataset, we compute the perplexity over 100 consecutive batches of 1024 tokens each.
In Table~\ref{tab:perp}, we compare the relaxed relative-threshold LAMP \eqref{eq:rrt_lamp} with its \emph{length-normalized} (LN) modification.
The idea of the latter is following: instead of using the same relative threshold $\tau$ for each row of the ``causal'' KQ matrix, irrespective of its length $n$, we try to increase the threshold for shorter rows to preserve the relative significance of the selected indices.
To test this idea, we scale the threshold as $\tau \sqrt{1024/n}$, where 1024 is the maximum context length the GPT-2 family of models was trained on.

First of all, the results in Table~\ref{tab:perp} show that the LAMP framework succeeds in decreasing the perplexity of purely low-precision inference using a small number of recomputed KQ inner products.
Moreover, the results obtained for CodeParrot demonstrate that the LN modification of relaxed LAMP can achieve the same perplexity as the original relaxed LAMP \eqref{eq:rrt_lamp} with fewer recomputations.
On the contrary, the results for GSM8k Math show that the LN variant can also perform worse than the original relaxed LAMP.
This suggests that length-normalized thresholding can be beneficial for LAMP, but empirical fine-tuning of the scaling law is necessary.

\begin{table}[t]
  \caption{Comparison of perplexities of full-precision inference, low-precision inference, relaxed LAMP \eqref{eq:rrt_lamp}, and its length-normalized modification applied to GPT-2 XL with $\mu = 4$ mantissa bits for KQ accumulation and validated on several datasets.}
  
  \label{tab:perp}
  \centering
  \begin{tabular}{lllll}
    \toprule
    Dataset & Method & Specification & Perplexity & Sparsity \\
    \midrule
    GSM8k Math & Full precision & N/A                              & 13.088 & 100\% \\
               & Low precision  & N/A                              & 13.161 & 0\%   \\
               & LAMP           & Relaxed \hfill($\tau = 0.03$)    & 13.091 & 5.8\% \\
               &                & Relaxed LN \hfill($\tau = 0.03$) & 13.094 & 4.5\% \\
               &                & Relaxed \hfill($\tau = 0.09$)    & 13.093 & 2.8\% \\
               &                & Relaxed LN \hfill($\tau = 0.09$) & 13.096 & 2.0\% \\
    \midrule
    WikiText-2 & Full precision & N/A                              & 17.411 & 100\% \\
               & Low precision  & N/A                              & 17.524 & 0\%   \\
               & LAMP           & Relaxed \hfill($\tau = 0.03$)    & 17.454 & 5.7\% \\
               &                & Relaxed LN \hfill($\tau = 0.03$) & 17.456 & 4.3\% \\
               &                & Relaxed \hfill($\tau = 0.09$)    & 17.458 & 2.7\% \\
               &                & Relaxed LN \hfill($\tau = 0.09$) & 17.459 & 1.9\% \\
    \midrule
    CodeParrot & Full precision & N/A                              & 3.174  & 100\% \\
               & Low precision  & N/A                              & 3.221  & 0\%   \\
               & LAMP           & Relaxed \hfill($\tau = 0.03$)    & 3.180  & 5.5\% \\
               &                & Relaxed LN \hfill($\tau = 0.03$) & 3.180  & 4.2\% \\
               &                & Relaxed \hfill($\tau = 0.09$)    & 3.182  & 2.6\% \\
               &                & Relaxed LN \hfill($\tau = 0.09$) & 3.183  & 1.8\% \\
    \bottomrule
  \end{tabular}
\end{table}

\subsection{Zero-shot multiple-choice task}
We additionally assess LAMP on a downstream task ARC-Easy\footnote{https://huggingface.co/datasets/allenai/ai2\_arc}.
The baseline accuracy achieved with full-precision GPT-2 XL inference is $0.561$, and it drops to $0.558$ with $\mu = 4$ mantissa bits for KQ accumulation.
The use of LAMP with $\tau = 0.09$ increases the accuracy to $0.560$ using $32\%$ of recomputations.
These results suggest that GPT-2 models are too small to be confident on zero-shot downstream tasks, which is likely why the accuracy drop is low and the recomputation rate is higher than in other experiments.

\end{document}